\documentclass[letterpaper, 10 pt, conference]{ieeeconf}  

\usepackage{times}
\usepackage{xcolor}
\usepackage{graphicx}
\usepackage{amsmath}
\usepackage{amssymb}
\usepackage{booktabs}
\usepackage{url}
\usepackage{multirow}
\usepackage{url}
\usepackage[space]{cite}
\usepackage[section]{placeins}
\usepackage[breaklinks=true,bookmarks=false]{hyperref}
\graphicspath{ {./images/} }
\usepackage[margin=0.792in]{geometry}

\IEEEoverridecommandlockouts                              

\title{\LARGE \bf
Retrieval and Localization with Observation Constraints}

\author{Yuhao Zhou$^{\dagger}$, Huanhuan Fan$^{\dagger}$, Shuang Gao$^{\dagger}$, Yuchen Yang, Xudong Zhang$^{\ast}$, Jijunnan Li, Yandong Guo
\thanks{${\dagger}$ Authors contributed equally to this work.}
\thanks{${\ast}$ indicates corresponding author. Contact: zhangxudong@oppo.com}
\thanks{All authors are with OPPO Research Institute, Shanghai, China.}
}

\begin{document}

\maketitle
\thispagestyle{empty}
\pagestyle{empty}

\begin{abstract}

Accurate visual re-localization is very critical to many artificial intelligence applications, such as augmented reality, virtual reality, robotics and autonomous driving. To accomplish this task, we propose an integrated visual re-localization method called RLOCS by combining image retrieval, semantic consistency and geometry verification to achieve accurate estimations. The localization pipeline is designed as a coarse-to-fine paradigm. In the retrieval part, we cascade the architecture of ResNet101-GeM-ArcFace and employ DBSCAN followed by spatial verification to obtain a better initial coarse pose. We design a module called observation constraints, which combines geometry information and semantic consistency for filtering outliers. Comprehensive experiments are conducted on open datasets, including retrieval on R-Oxford5k and R-Paris6k, semantic segmentation on Cityscapes, localization on Aachen Day-Night and InLoc. By creatively modifying separate modules in the total pipeline, our method achieves many performance improvements on the challenging localization benchmarks.
\end{abstract}

\section{INTRODUCTION}

Visual Localization serves as the fundamental capability of numerous vision applications, including augmented reality, intelligent robotics and autonomous driving navigation \cite{Lim2013Real, Castle2008Video}. This approach's core task is to estimate the 6-degrees of freedom (DoF), i.e., the position and orientation of a query RGB image in a known 3-Dimensional (3D) coordinate environment.

The presentation of the environment can be a map reconstructed by Structure From Motion (SFM) \cite{wang2015adaptive, schonberger2016structure, ullman1979interpretation}, a database of images \cite{sattler2017large,taira2018inloc}, or even regression Convolutional Neural Network (CNN) \cite{kendall2015posenet}.
In detail, the SFM based map is typically used to describe the position of landmarks \cite{fuentes2015visual, valgren2010sift}, i.e., 3D points and structures in the environments, which are pre-collected and extracted from the database images. During localization, correspondences between 2D keypoints and 3D landmarks are established to recover the query image's 6-DoF pose using Perspective-n-Point (PnP) \cite{gao2003complete} within a RANSAC loop \cite{fischler1981random, chum2003locally}. To avoid costly timing on searching and matching in irrelevant mapping areas, image retrieval is used to select the most relevant database images \cite{taira2018inloc, sarlin2019coarse}. 
Local feature matching is then established between the query image and the area defined by retrieved database images.

Since the correspondences between query and database images need to be established in visual localization tasks, environment changes, such as weather, illumination, or seasonal changes, present critical challenges for local feature descriptors.

The traditional local features and descriptors, e.g., SIFT \cite{lowe2004distinctive}, BRIEF \cite{calonder2010brief}, or ORB \cite{rublee2011orb}, which have been carefully designed for uniform intensity changes and slight variations of viewpoints, were shown to be highly sensitive to massive changes in lighting and seasonal conditions.
An attempt at overcoming conditions-changing is training convolutional neural networks (CNNs) to produce more robust feature descriptors \cite{yi2016lift, detone2018superpoint, revaud2019r2d2, dusmanu2019d2}, instead of using handcraft features.
Although CNNs are shown to have great improvements compared to SIFT and other handcrafted features, they were not designed to handle all the types of variations described above.
As feature detectors and descriptors are less repeatable and reliable, localization pipelines then struggle to find enough query-to-database correspondings to recover successful pose estimation. Therefore, developing more robust localization pipelines that work well across a wider range of environmental conditions is desirable.

In this paper, we present a localization pipeline, Retrieval and Localization with Observation ConstraintS (RLOCS), which utilizes image retrieval to acquire coarse initial localization poses and combines geometric and semantic information to refine the localization results.

The core idea of RLOCS is to employ a natural coarse-to-fine strategy for recovering 6-DoF poses of query images in the related pre-built SFM model.
In detail, RLOCS leverages both global descriptors for image-retrievals and local features for semantic-matching to establish a localization pipeline.
We show that RLOCS, using CNN-based image retrieval method and hybrid local descriptors, enables robustness and reliable results under many challenging conditions. Our global descriptors outperform most previous results in the retrieval task, and the learning-based local features improve the accuracy of pose estimation.

Meanwhile, inspired by previous work on pose verification via observation \cite{taira2018inloc, toft2018semantic}, we propose a 6-DoF pose optimization method based on observation constraints of the query image. 
The pose optimization starts with a standard PnP within a RANSAC loop and obtains an initial pose estimation.
The 3D points with locations and descriptors are then collected using the initial pose and observation constraints.
Consequently, matches between 2D points from the query image and 3D points from the SFM model are established to refine the initial pose. 
More details of pose optimization methods are discussed in the following sections.

In summary, our contributions include the following key enhancements on the retrieval-based visual localization pipeline:

1. A better retrieval CNN is proposed, followed by a clustering and a spatial verification method known as re-rank. We regress the coarse localization initial pose by a 2D-2D matching module. 

2. A coarse-to-fine two-stage localization pipeline using observation constraints as the back-end fine-tuning
optimization method is utilized, which contains geometry attributes
and semantic information.

3. An efficient and accurate semantic segmentation CNN structure is adopted and optimized to achieve better semantic precision, which finally benefits the localization accuracy.

\section{Related Work}

In this section, we will discuss recent approaches related to different components of our works: large-scale image retrieval tasks, semantic segmentation and visual localization. 

\textbf{Large-scale Image Retrieval.} As a classical problem in computer vision,  large-scale image retrieval has been widely analyzed in the past decades. For statistically quantifying the performance of different methods, several standard datasets have been published and widely used, e.g., R-Oxford5k, R-Paris6k \cite{radenovic2018revisiting}, and Google Landmarks dataset v1, v2 \cite{noh2017large, weyand2020google}. The diversity of environments and challenges of illumination changes are presented in these large datasets. In the retrieval tasks, approaches can be divided into two categories, local features and global features. As to the local features, some methods aggregate local features into global descriptors, such as VLAD \cite{jegou2010aggregating} and FV \cite{jegou2011aggregating}, while others form a feature model for searching and query, like BOW \cite{sivic2003video} and other related methods \cite{philbin2007object, philbin2008lost}. For the global features, with the rapid development of deep learning, CNN-based methods outperform most of the hand-crafted methods with higher recall and accuracy. Using these CNN-based global descriptors, \cite{arandjelovic2016netvlad} makes use of deep local features as a drop-in replacement for hand-crafted features in conventional aggregation such as VLAD. \cite{noh2017large} relies on CNN to produce attentive local features and regresses them into global indexes. \cite{cao2020unifying} unifies local and global descriptors into a single CNN with generalized mean pooling and attention selection modules.
 
\textbf{Semantic Segmentation.} Plenty of researchers have analyzed many strategies and structures on semantic segmentation CNNs in recent years. \cite{Chen2017Rethinking} adopts dilation convolutional layers to achieve larger receptive area, while \cite{2017Large} utilizes large kernel convolutional layers. \cite{2018BiSeNet} relies on the spatial attention module to enlarge the structure information, while \cite{2018Dual} adaptively integrates local and global features by spatial-wise and channel-wise attention modules. \cite{2016Deep} exploits shortcuts between multiple layers to avoid degradation problems, and \cite{2016Densely} proves the better performance with the help of shortcut connections between layers in a feed-forward fashion. Unlike previous works, we integrated several modules to lift the semantic segmentation precision and analyzed the positive inference on visual localization brought by the semantic information.

\textbf{Visual Localization.} Among all the image-based localization approaches, structure-based ones are one of the most extensively discussed methods, which extract 2D key-points of a query image and find matching relationships between 3D points constructed by SFM \cite{Svarm2016City}. \cite{sattler2016efficient} expands from original 2D-to-3D matches to 3D-to-2D matches to realize a better localization performance. \cite{toft2018semantic} adds semantic information to help filter some outliers of key-points matching. \cite{sarlin2019coarse} merges local features extractor and global one into a single CNN to realize higher efficiency and robustness. \cite{zhang2020reference} relies on renderings of a 3D model to produce better local features matching and seeks a better final estimated pose iteratively.

\section{Visual Re-Localization Pipeline}

\begin{figure*}
	\includegraphics[width=16cm]{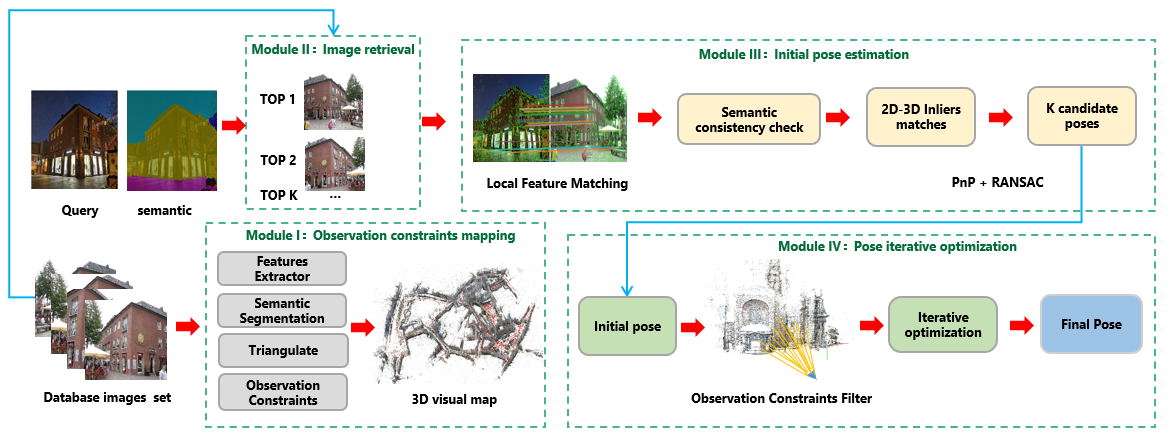}
	\centering
	\caption{
		\textbf{ Visual Re-localization Pipeline}. The input data contains both the database images set for mapping and a query image for Re-localization. Module I is the construction of observation constraints map, including SuperPoint and R2D2 local feature extraction and matching, semantic segmentation, triangulation of matched image pairs, BA optimization and calculation of observation constraints. The output of Module I is a semantic 3D map with observation constraints. Module II is the image retrieval part, returning ${K}$ images from database images set, which are most similar to the query image. Module III is to obtain a rough pose. The process includes the 2D-2D matching between query and candidate images, semantic consistency verification, pose clustering and solving PnP using 2D-3D inlier matches. Module IV is the iterative optimization.}
	\label{fig:pipeline}
\end{figure*}

Our Re-localization method, named RLOCS, consists of four parts: mapping with observation constraints, image retrieval, initial pose estimation and iterative pose optimization. Fig.\ref{fig:pipeline} illustrates the pipeline based on a standard retrieval-based framework in \cite{irschara2009structure}. 

\textbf{Observation constraints mapping.} We run COLMAP \cite{schonberger2016structure}, a superior SFM algorithm to reconstruct a 3D model and camera poses of the database images captured at the target scene. After triangulation, 3D map points will be produced based on epipolar geometry theory. More semantic and geometric information, regarded as observation measurements, can be calculated and tagged on every 3D point for checking and filtering in the visual localization process.

\textbf{Image retrieval.} The landmark image search is performed by matching the query with the database images using the global descriptor calculated by CNN described detailedly in \ref{retrieval section}. Through this retrieval scheme, a fixed number of similar database images are collected. DBSCAN \cite{birant2007st} and a re-rank \cite{cao2020unifying} procedure are adopted to fine-tune the retrieval results. 

\textbf{Initial pose estimation.} For every candidate, the semantic consistency check is applied to filter out inaccurate keypoints matches. The filtered 2D-3D matches will determine the 6-DoF camera pose by solving a PnP geometric consistency check inside a RANSAC loop. This algorithm is called Feature Match(FM)-PnP module in our method.

\textbf{Pose iterative optimization.} Given the coarse 6-DoF pose, more geometric and semantic observation constraints are applied to select the 3D points. After selection, K-Nearest Neighbor (KNN) matching enables us to get the 2D-3D matches. Such process is conducted iteratively to make the final estimated pose closer to the ground truth.

In the total localization pipeline, our contributions can be concluded into three main parts, i.e., image retrieval, semantic segmentation, and observation constraints.

\begin{figure}
	\includegraphics[width=8cm]{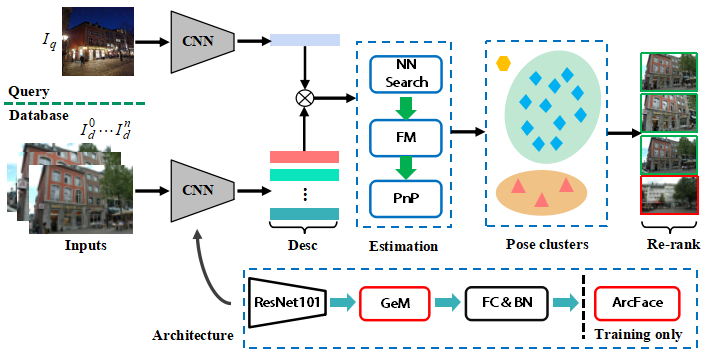}
	\centering
	\caption{
		\textbf{Illustration of retrieval strategy in RLOCS.} Firstly, global features are extracted from the cascaded architecture of ResNet101-GeM-ArcFace. These features are then fed into NN search in database images. For every retrieved candidate from the database, an FM-PnP pipeline is performed for solving coarse 6-DoF poses of the query image. And DBSCAN clustering method, followed by the re-rank procedure, is applied for seeking more accurate retrievals results.
	}
	\label{fig:retrievalmodel}
\end{figure}

\subsection{Image Retrieval}
\label{retrieval section}

In RLOCS, a coarse retrieval task is first performed by matching the query with the database images using global descriptors in both SFM mapping and localization pipeline.
We leverage recent improvements in global feature designs, such as a Generalized Mean pooling (GeM) \cite{radenovic2018fine} layer and ArcFace \cite{deng2019arcface} loss to generate effectively aggregated global descriptors. The strategy of global retrieval is illustrated in Fig. \ref{fig:retrievalmodel}. 

Given a query image, a basic backbone CNN is first adopted to obtain the feature map, representing deep activations.
GeM pooling is applied to weigh each feature map's contributions and aggregate the activations into a fixed-length global descriptor. 
In the work of \cite{radenovic2018fine}, GeM is shown superior performance than other pooling methods, such as regional max-pooling (R-Mac) \cite{lin2018regional} and sum-pooled convolutional features (SPoC) \cite{babenko2015aggregating}. The defination of GeM can be described as
\begin{equation} \label{eq:gem}
f_{c}^{g} = (\frac{1}{|X_c|}\sum\limits_{x\in{X_c}}x^{p})^{\frac{1}{p}},
\end{equation}
where $x$ is the feature at each location of $X_c$, which is extracted from the backbone and $p$ denotes the generalized mean power parameter.
Note that the $p$ of GeM pooling is set to 3.0 and fixed during our training process\cite{radenovic2018fine}.
Dimension reduction is then adopted behind the GeM pooling, adding a fully connected layer cascaded with one-dimensional Batch Normalization, which is crucial to alleviate the risk of over-fitting and reduce the dimensional noise.

For better performance on global feature learning, we utilize the ArcFace\cite{deng2019arcface} margin-based loss as the training components, which has achieved impressive results by including smaller intra-class variance in face recognition.
The ArcFace is defined as 
\begin{equation} \label{eq:arcface}
L = -\frac{1}{N}\sum_{i=1}^N\log\frac{e^{W_{y_{i}}^{T}x_{i}+b_{y_{i}}}}{\sum_{j=1}^{n}e^{W_j^{T}x_i+b_j}},
\end{equation}
where $x_i\in\Re^d$ denotes the deep features of the $i$-th sample in $d$ dimensions, belonging to the $y_{i}$-th class.  $W_{y_{i}}$ denotes the weights term of $y_{i}$-th class. $W_j\in\Re^d$ and $b_j\in\Re^d$ are the $j$-th column of the weights and the bias term, respectively.
In this work, we follow \cite{deng2019arcface} and train our retrieval model with image-level annotations on the Google Landmarks dataset v2 \cite{weyand2020google}.
For evaluation components, we adopt the learned fixed-length global descriptor following by an L2-normalization and Principal Component Analysis (PCA) process for all the query and database images.

Secondly, a KNN search is performed by matching query images with the candidates using our global descriptors. However, each landmark category in the database may contain diverse samples, such as variation of viewpoints and illumination.
These query images are tough to identify only using context-level global features. Therefore we employ a back-end discriminative clustering method to exploit the 6-DoF poses from the database.
In detail, an FM-PnP strategy is firstly performed for initial 6-DoF poses between query images and the retrieved top-$k$ images from the database.
These poses are then clustered based on the inliers and distances using DBSCAN \cite{birant2007st}, following by a PnP spatial verification algorithm \cite{cao2020unifying}.

\subsection{Semantic Segmentation}
\label{section_semantic}

\begin{figure}
	\includegraphics[width=8cm]{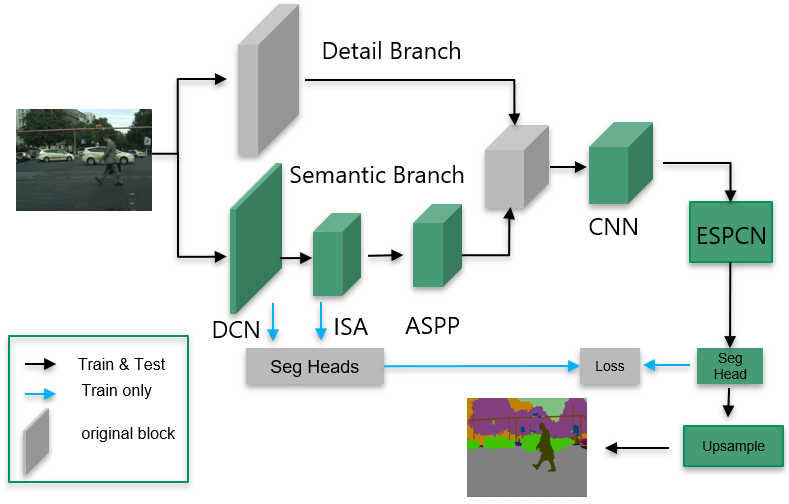}
	\centering
	\caption{
		\textbf{Illustration of proposed semantic segmentation architecture.} Feature maps are separated into semantic and detail branches. DCN, ISA and ASPP are applied to acquire a larger receptive area, gaining more global and shape information in the semantic branch. And ESPCN is used to recover stable and uniform semantic masks.
	}
	\label{fig:Bilateral_attention_semantic}
\end{figure}


Considering that semantic segmentation is relatively stable under illuminational changes, we take advantage of semantic segmentation algorithms as one of the observation constraints during the mapping and localization process, to improve the pose estimations' accuracy.

For pixel-level semantic segmentation tasks, efficiency and accuracy are both significant. In the work of \cite{yu2020bisenet}, a state-of-the-art network, BiSeNet-V2, meets both the high-speed and accuracy demands in our localization pipeline. 

Two branches, i.e., detail branch and semantic branch, are inherited from \cite{yu2020bisenet}. Detail branch is meant to produce low-level features with shallow CNN layers. To enhance the receptive field and capture rich contextual information in the semantic branch, we adopt the Interlaced Sparse Self-Attention (ISA) \cite{huang2019interlaced} spatial attention mechanism in the semantic branch to enhance the receptive field. And the additional Atrous Spatial Pyramid Pooling (ASPP) \cite{chen2018encoder} module together with ISA helps to solve the multi-scale problems for objects. 
We use Deformable Convolutional Networks (DCN)\cite{dai2017deformable} module to modify the semantic branch backbone to make the branch paying more attention to the shapes of different objects.
Afterward, the aggregation layer manages to merge both detail and semantic branch into a feature map.  


With the help of Efficient Sub-Pixel Convolutional Neural Network (ESPCN), the feature map can be upsampled by a factor of 4. Ending with an additional 3x3 convolution layer, which is cascaded with a bilinear upsampling layer of factor 2, ensures us to get a stable and uniform semantic mask.

\subsection{Observation Constraints and Pose Iterative Optimization}

The original 3D point clouds reconstructed by COLMAP \cite{schonberger2016structure} only contain the primary attributes, such as position coordinates and color. In this paper, we add more attributes named observation constraints based on the idea of semantic consistency \cite{toft2018semantic}, including semantic and geometric constraints. The additional attributes of every 3D point will guarantee the filtering process before 2D-3D matching in the localization part. As to local features, we inherit both Super-Point \cite{detone2018superpoint} and R2D2 \cite{revaud2019r2d2} to extract 2D points and descriptors.

\begin{figure}
	\includegraphics[width=8cm]{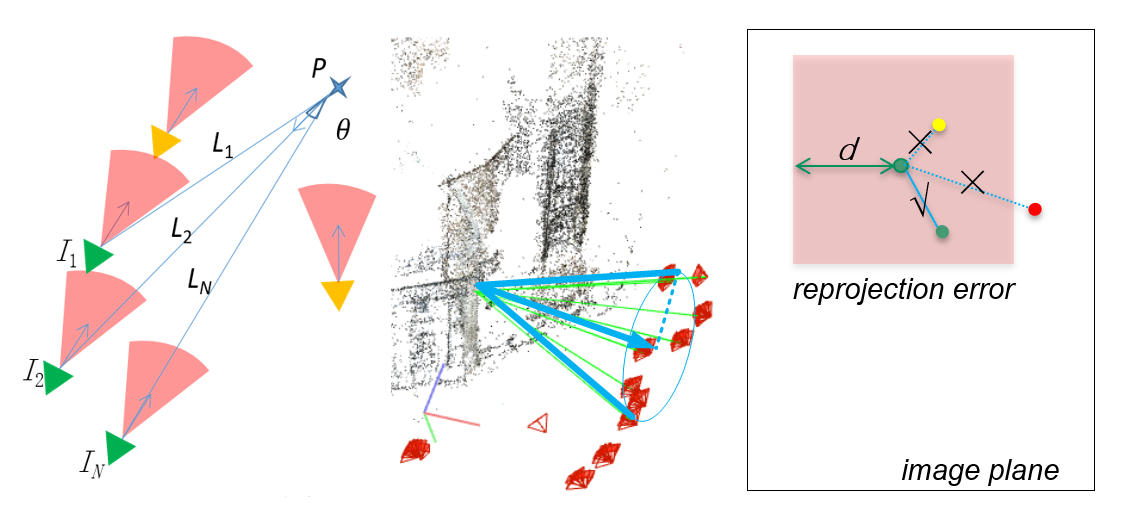}
	\centering
	\caption{
		\textbf{Illustration of Observation Constraints}. The left figure is the schematic diagram and ${N}$ is the number of images associated with the 3D point ${P}$. ${L_i}$ is the distance between point ${P}$ and the optical center of the camera ${I_i}$. The middle figure shows a 3D point visual field which is a cone area in the map. The right one shows how the reprojection error and the semantic label function as observation constraints. ${d}$ is the error threshold and points' color represents the semantic labels.
	}
	\label{fig:OC}
\end{figure}

The additional information on each point consists of maximum visible distance, mean visible direction, maximum visible angle, semantic label and reprojection error. 
During the SFM process, we backtrace each 3D point to find the 2D feature points and the corresponding images that participate in the triangulation. Supposing the 3D point $P$ has $N$ track elements, the corresponding images are denoted as ${I_1},{I_2},{...},{I_N}$. As the schematic diagram illustrated in Fig.\ref{fig:OC}, the visible field of a 3D point is the cone area. Whether a query image can see the 3D point depends on whether its pose is in the visible field.

The maximum visible distance ${L}$, the mean visible direction ${\vec n}$ and the maximum visible angle ${\theta}$ can be formulated as follows:
\begin{equation} \label{eq:oc_1}
{L}=\mathop{\max_{i}}\|X-C_i\|_{2} \quad{i\in[1...N]},
\end{equation}
\begin{equation} \label{eq:oc_2}
\vec n=\frac{1}{N}\sum_{i=1}^{N}{\frac{{\overrightarrow{C_i}-\overrightarrow{X}}}{\sqrt{{\|C_i-X\|}^{2}}}},
\end{equation}
\begin{equation} \label{eq:oc_3}
\theta=2\mathop{\max_{i}}({arccos}({\vec n}{\cdot}{\frac{\overrightarrow{C_i}-\overrightarrow{X}}{\sqrt{{\|C_i-X\|}^{2}}}})) \quad{i\in[1...N]},
\end{equation}
where ${X}$ denotes position coordinates of point ${P}$, and ${C_i}$ denotes the camera optical center position of image ${I_i}$. $\overrightarrow{X}$ and $\overrightarrow{C_i}$ are the vector representations of ${X}$ and ${C_i}$ respectively. The visible field of point ${P}$ is a cone area that is determined by  ${L}$, ${\vec n}$ and ${\theta}$. The ${L}$ is the cone bus, which means the farthest distance where the point ${P}$ can be seen. The ${\vec n}$ represents the average value of the direction from point ${P}$ to the corresponding camera optical center ${C_i}$ as the normal of ${P}$. The ${\theta}$ equals two times the maximum angle between the normal ${\vec n}$ and all of the ${\overrightarrow{PI_i}}$. We adopt a voting strategy and determine the 3D semantic label with the largest occurrence frequency of correlated 2D points. 

	
 Each initial pose is obtained by 2D-3D inliers and PnP process described as Module III in Fig.\ref{fig:pipeline}.	
Then the pose iterative optimization is performed by using observation constraints. The specific steps are listed as follows: 
\begin{enumerate}
	\item Select 3D points with initial pose and the visible field determined by ${L}$, ${\vec n}$, ${\theta}$.
	
	\item Produce global 2D-3D matches by KNN and semantic consistency check.
	\item Filter outliers of 2D-3D matches within a certain threshold of re-projection error.
	\item Compute the 6-DoF camera pose by solving a PnP algorithm inside a RANSAC loop.
	\item Update the pose of the query if the convergence condition is fulfilled.
\end{enumerate}

The convergence condition is determined by the uncertainty quantification of the iterative pose. Based on the Monte Carlo Sampling\cite{sobol1994primer}, we first randomly sample $k$\% (e.g., $k=30,50,70$) sub-matches from all the global 2D-3D matches. The sub-poses are then calculated from sub-matches using the PnP algorithm. 
The standard deviation between all the sub-poses and current pose is defined as the sampling uncertainty. It is supposed to get smaller during the iteration loops. Otherwise, the optimization stops and the final prediction is produced.




\section{Experimental Evaluation}
\label{section: experiment}

The performance of whole visual re-localization is discussed in the following part. Experiments are executed on one NVIDIA Tesla V100 with the CUDA 10.0 and Intel(R) Xeon(R) Gold 6142 CPU @ 2.60GHz. On average, RLOCS consumes 198ms per frame.

\subsection{Ablation Study}

\textbf{Image Retrieval.} For large-scale image retrieval tasks, we conduct our evaluation of the proposed ResNet101-GeM-ArcFace pipeline on \textit{R-Oxford5k} and \textit{R-Paris6k} datasets. The Google Landmarks dataset v2 \cite{weyand2020google} is the training set. Table \ref{table1} shows that our method outperforms some of the state-of-art retrieval methods statistically. 

\begin{table}[htbp]
\centering
\caption{retrieval result(${mAP}$) on \textit{R-Oxford5k} and \textit{R-Paris6k} with both medium and hard evaluation protocols.}
\label{table1} 
\renewcommand\arraystretch{1.6}
\begin{tabular}{ccccc}
\toprule
\multirow{2}*{Methods} & \multicolumn{2}{c}{\textit{R-Oxford5k}} & \multicolumn{2}{c}{ \textit{R-Paris6k}}  \\
~ & Medium & Hard & Medium & Hard \\
\midrule
{NetVLAD\cite{arandjelovic2016netvlad}} & 63.5 &- & 73.5 & - \\  
ResNet101-RMAC\cite{gordo2017end} & 60.9 & 32.4 & 78.9 & 59.4 \\  
ResNet101-GeM-AP\cite{revaud2019learning} & 67.5 & 42.8 & 80.1 & 60.5 \\  
DELG\cite{cao2020unifying} & 69.7 & 45.1 & 81.6 & 63.4 \\  
\textbf{Ours} & \textbf{72.5} & \textbf{55.9} & \textbf{85.8} & \textbf{71.8} \\ 
\bottomrule
\end{tabular}
\end{table}

While DELG utilizes ResNet50 to extract low-level feature maps, ours relies on larger ResNet101 to acquire more features to get better retrieval accuracy. As GeM pooling layers are meant to maintain more information than original max-pooling layers during the CNN inference, RLOCS using GeM as a pooling layer outperforms the ResNet101-RMAC whose pooling layers are derived from the max-pooling layers. Evidence \cite{deng2019arcface} also indicates that margin-based loss will efficiently escalate the discriminative power between different classes. Thus RLOCS outperforms both \cite{gordo2017end} and \cite{revaud2019learning} on these test datasets.

\textbf{Semantic Segmentation.} Cityscapes dataset \cite{cordts2016cityscapes} is a widely used semantic segmentation benchmark that contains urban street scenes. The final results on its test sets show that the combination of ASPP, DCN, ESPCN, and ISA modules can a achieve better mean Intersection of Union (mIoU) with reasonable speed. Statistics show that progressively adding these modules results in gradually increasing accuracy.
Simultaneously, due to the additional calculation introduced by these modules in the feature extraction parts, the inference time increases as shown in table \ref{table2}. Compared with the state-of-the-art real-time semantic segmentation model on Cityscapes test dataset, we achieve better accuracy and faster speed, demonstrated in Table \ref{table2_2}.

\begin{table}[htbp]
\centering
\caption{ablation study of proposed modules on \textit{Cityscapes} dataset}
\label{table2}
\renewcommand\arraystretch{1.6}
\setlength{\tabcolsep}{1.2mm}
\begin{tabular}{cccccccc}
\toprule
ASPP & DCN & ESPCN & ISA & val mIoU($\%$) & test mIoU($\%$) & Time(ms) \\
\midrule
\checkmark&- & -& -& 74.0& 71.2 & \textbf{10.4} \\ 
\checkmark&\checkmark & -& -& 75.0& 72.4 & 11.0 \\ 
\checkmark&\checkmark & \checkmark& -& 77.2& 75.9& 12.8 \\ 
\checkmark&\checkmark & \checkmark&  \checkmark& \textbf{78.5} & \textbf{76.5}  & 14.1  \\
\bottomrule
\end{tabular}
\end{table}

\begin{table}[htbp]
\centering
\caption{semantic segmentation results on \textit{Cityscapes} testset}
\label{table2_2}
\renewcommand\arraystretch{1.6}
\setlength{\tabcolsep}{1.2mm}
\begin{tabular}{cccc}
\toprule
Methods & val mIoU($\%$) & test mIoU($\%$) & Time(ms) \\
\midrule
 BiSeNet V2-Large\cite{yu2020bisenet}& 75.8 & 75.3& 21.1 \\ 
 SwiftNetRN-18\cite{orsic2019defense}& -& 75.5& 25.0 \\ 
 U-HarDNet-70\cite{chao2019hardnet}& 75.4& 75.9& 18.8 \\ 
\textbf{Ours}& \textbf{78.5} & \textbf{76.5}  & \textbf{14.1}  \\
\bottomrule
\end{tabular}
\end{table}

\subsection{Localization Performance}

We evaluate our Re-localization method using the online benchmark, which calculates the percentage of query images within three different thresholds of rotation and translation error. Two types of datasets, Aachen Day-Night \cite{sattler2018benchmarking} and InLoc \cite{taira2018inloc}, including indoor and outdoor scenes, are used for validation. 

The improvement in localization accuracy brought by our image retrieval method proves the effectiveness of our scheme. We conduct the experiments by changing different retrieval schemes followed by the same Re-localization pipeline on Aachen Day-Night dataset. Compared with DELG, one of the state-of-the-art retrieval methods illustrated in Table \ref{table1}, after applying the same DBSCAN clustering scheme, our method still performs better. The results are shown in Table \ref{table3}. Our method has an absolute advantage on the night dataset, while on day datasets, ours is also the best one under the largest threshold. As retrieval gives a coarse initial pose, a more accurate and robust method will relieve much pressure on backend localization procedures and has more significant benefits on the larger accuracy threshold. 

\begin{table}[htbp]
\centering
\caption{re-localization accuracy on \textit{Aachen Day-Night v1.1} using different retrieval schemes}
\label{table3}
\renewcommand\arraystretch{1.6}
\begin{tabular}{ccc}
\toprule
\multirow{2}*{Methods}  & \multicolumn{2}{c}{Accuracy(0.25m, 2°)/(0.5m, 5°)/(5m, 10°)}  \\
~  &  {Day} &  {Night}\\
\midrule
DELG  & 88.8 / \textbf{95.9} / 98.8 & 69.6 / 84.8 / 94.8 \\ 
DELG + DBSCAN  & \textbf{89.2} / 95.5 / 98.5 & 72.3 / 88.0 / \textbf{98.4}  \\ 
\textbf{Ours}  & 88.8 / 95.4  / \textbf{99.0} & \textbf{74.3} / \textbf{90.1} / \textbf{98.4} \\ 
\bottomrule
\end{tabular}
\end{table}

Considering semantic segmentation is a part of the observation constraints for the pose optimization, we validate their inference to the localization accuracy and other geometric attributes. We adopt the retrieval results in Table \ref{table3} as the initial poses to be optimized by using observation constraints on the Aachen dataset. 
On the InLoc dataset, we validate the influence introduced by observation constraints with and without semantic segmentation, as shown in Table \ref{table5}. We found that almost all the accuracy is improved under the observation constraints optimization, and further improved under the one with semantic information. 



\begin{table}[htbp]
\centering
\caption{re-localization accuracy on \textit{InLoc} with observation constraints}
\label{table5}
\renewcommand\arraystretch{1.6}
\begin{tabular}{ccc}
\toprule
\multirow{2}*{Methods}  & \multicolumn{2}{c}{Accuracy(0.25m, 2°)/(0.5m, 5°)/(5m, 10°)}  \\
~  &  {duc1} &  {duc2}\\
\midrule
BaseLine  & 41.9 / 68.2 / 84.3 & 50.4 / 76.3 / 80.2 \\ 
+OC(w/o semantic)  & \textbf{47.0} / 68.7 / \textbf{84.8} & 57.3 / 76.3 / \textbf{80.9} \\ 
+OC(w/ semantic)  & \textbf{47.0} / \textbf{71.2} / \textbf{84.8} & \textbf{58.8} / \textbf{77.9} / \textbf{80.9} \\ 
\bottomrule
\end{tabular}
\end{table}

Totally, we compare our proposed pipeline with some existing state-of-the-art approaches at the Long-Term Visual Localization benchmark 2020 \cite{Lim2013Real}. We capture the latest results of various typical approaches from \url{visuallocalization.net/benchmark/} and show in Table \ref{table6}. Statistically, our methods show better accuracy compared to some of the localization methods on Aachen Day-Night and InLoc datasets, especially on harder night subsets and indoor datasets, including many occlusions. Alabtion study on observation constraints has been conducted in \ref{table6} to show the improvements brought by our proposed method.

\begin{table}[htbp]
	\centering
	\caption{evaluation of state-of-the-art approaches on \textit{Aachen Day-Night v1.0, v1.1} and \textit{InLoc}}
	\label{table6}
	\renewcommand\arraystretch{1.4}
	\setlength{\tabcolsep}{1mm}{
		\begin{tabular}{ccc}
			\toprule
			\multirow{2}*{\textit{Aachen Day-Night v1.0}} & \textit{day} & \textit{night} \\
			~ & \multicolumn{2}{c}{{(0.25m, 2°)/(0.5m, 5°)/(5m, 10°)}} \\
			\midrule
			{Active Search v1.1\cite{sattler2016efficient}} & 57.3 / 83.7 / 96.6 & 19.4 / 30.6 / 43.9 \\ 
			{NetVLAD + D2-Net\cite{dusmanu2019d2}} & 84.8 / 92.6 / 97.5 & 84.7 / 90.8 / 96.9 \\ 
			{DenseVLAD + D2-Net\cite{dusmanu2019d2}} & 83.1 / 90.9 / 95.5 & 74.5 / 85.7 / 90.8 \\ 
			{KAPTURE-R2D2-APGeM\cite{humenberger2020robust}} & 88.7 / 95.8 / 98.8 & 81.6 / 88.8 / 96.9 \\ 
			{SuperPoint + SuperGlue\cite{sarlin2020superglue}} & \textbf{89.6} / \textbf{95.4} /98.8 & \textbf{86.7} / \textbf{93.9} / \textbf{100.0} \\
			\textbf{Ours(w/o OC)} & 85.7 / 93.7 / 98.9 & 81.6 / 91.8 / 100.0 \\
			\textbf{Ours(w/ OC)} & 88.8 / \textbf{95.4} / \textbf{99.0} & 85.7 / \textbf{93.9} / \textbf{100.0} \\ 
			\hline
			\multirow{2}*{\textit{Aachen Day-Night v1.1}} & \textit{day} & \textit{night} \\
			~ & \multicolumn{2}{c}{{(0.25m, 2°)/(0.5m, 5°)/(5m, 10°)}} \\
			\hline
			{Isrf-5k-o2s\cite{melekhov2020image}} & 87.1 / 94.7 / 98.3  & 74.3 / 86.9 / 97.4 \\
			{LISRD+SuperPoint\cite{pautrat2020online}} & - / - / - & 72.3 / 86.4 / 97.4 \\
			{KAPTURE-R2D2-APGeM\cite{humenberger2020robust}} & \textbf{90.0} / \textbf{96.2} / \textbf{99.5} & 72.3 / 86.4 / 97.9 \\
			{SuperPoint + SuperGlue\cite{sarlin2020superglue}} & 89.8 / 68.7 / 80.8 & \textbf{77.0} / \textbf{90.6} / \textbf{100.0} \\
			\textbf{Ours(w/o OC)} & 85.7 / 93.7  / 98.9 & 74.3 / 90.1 / 98.4 \\
			\textbf{Ours(w/ OC)} & 88.8 / 95.4  / 99.0 & 74.3 / \textbf{90.6} / 98.4 \\ 
			\hline
			\multirow{2}*{\textit{InLoc}} & \textit{duc1} & \textit{duc2} \\
			~ & \multicolumn{2}{c}{{(0.25m, 10°)/(0.5m, 10°)/(5m, 10°)}} \\
			\hline
			{HF-Net\cite{sarlin2019coarse}} & 39.9 / 55.6 / 67.2 & 37.4 / 57.3 / 70.2 \\
			{Isrf-5k-o2s\cite{melekhov2020image}} & 39.4 / 58.1 / 70.2 & 41.2 / 61.1 / 69.5 \\
			{Sparse-NCNet\cite{rocco2018neighbourhood}} & 47.0 / 67.2 / 79.8 & 43.5 / 64.9 / 80.2 \\ 
			{D2-Net\cite{dusmanu2019d2}} & 42.9 / 63.1 / 75.3 & 40.5 / 61.8 / 77.9 \\ 
			{KAPTURE-R2D2-FUSION\cite{humenberger2020robust}} & 41.4 / 60.1 /  73.7 & 47.3 / 67.2 / 73.3\\
			{SuperPoint + SuperGlue\cite{sarlin2020superglue}} & \textbf{49.0} / 68.7 / 80.8 & 53.4 / 77.1 / \textbf{82.4} \\ 
			\textbf{Ours(w/o OC)} & 41.9 / 68.2 / 84.3 & 50.4 / 76.3 / 80.2 \\
			\textbf{Ours(w/ OC)} & 47.0 / \textbf{71.2} / \textbf{84.8} & \textbf{58.8} / \textbf{77.9} / 80.9 \\ 
			\bottomrule
	\end{tabular}}
	
\end{table}

\section{CONCLUSIONS}

In conclusion, an integrated visual re-localization method named RLOCS is proposed. A more accurate and robust retrieval CNN is designed, and coarse initial localization poses are produced by DBSCAN clustering, spatial verification and 2D-2D matching. Furthermore, an optimization scheme called observation constraints containing semantic segmentation and other geometry attributes is adopted to iteratively fine-tune the poses. Abundant experiments are conducted on Aachen Day-Night and InLoc datasets to prove our method's effectiveness, with comparisons to some state-of-the-art visual localization methods. The entire pipeline has great expansibility and potential, such as further improving the image retrieval CNN or semantic CNN and including more geometric hints as additional constraints, like depths or normals of every 3D point.

\newpage

\bibliographystyle{ieeetr}
\bibliography{rlocs_bib}

\end{document}